\documentclass[letterpaper]{article}
\usepackage{aaai}
\usepackage{times}
\usepackage{helvet}
\usepackage{courier}
\usepackage{graphicx}
\usepackage[square,sort,comma,numbers]{natbib}
\usepackage{booktabs}
\title{Topic Detection and Summarization of User Reviews}
\author{Pengyuan Li\textsuperscript{\rm 1}, Lei Huang\textsuperscript{\rm 2}, Guang-jie Ren\textsuperscript{\rm 2}\\
\textsuperscript{\rm 1}Department of Computer and Information Sciences, \\University of Delaware, Newark, Delaware, USA\\
\textsuperscript{\rm 2}IBM Almaden Research, San Jose, USA\\
pengyuan@udel.edu,
lei.huang1@ibm.com, gren@us.ibm.com}

\begin{document}
\maketitle

\begin{abstract}
\begin{quote}
A massive amount of reviews are generated daily from various platforms. 
It is impossible for people to read through tons of reviews and to obtain useful information. 
Automatic summarizing customer reviews thus is important for identifying and extracting the essential information to help users to obtain the gist of the data. 
However, as customer reviews are typically short, informal, and multifaceted, it is extremely challenging to generate topic-wise summarization.
While there are several studies aims to solve this issue, they are heuristic methods that are developed only utilizing customer reviews. 
Unlike existing method, we propose an effective new summarization method by analyzing both reviews and summaries.
To do that, we first segment reviews and summaries into individual sentiments. 
As the sentiments are typically short, we combine sentiments talking about the same aspect into a single document and apply topic modeling method to identify hidden topics among customer reviews and summaries. 
Sentiment analysis is employed to distinguish positive and negative opinions among each detected topic. 
A classifier is also introduced to distinguish the writing pattern of summaries and that of customer reviews. 
Finally, sentiments are selected to generate the summarization based on their topic relevance, sentiment analysis score and the writing pattern. 
To test our method, a new dataset comprising product reviews and summaries about 1028 products are collected from Amazon and CNET. Experimental results show the effectiveness of our method compared with other methods. 
\end{quote}
\end{abstract}

\section{Introduction}
The number of customer reviews from various platforms grows rapidly nowadays.
It is impossible for people to read through tons of reviews and to obtain useful information. Automatic summarizing customer reviews thus is important for identifying and extracting the essential information to help users to understand or to get the gist of the data.
By providing a summarization of previous product reviews, customers can easily understand the features of products and sellers can learn the actual needs from customers' feedback. 
Many studies have been focused on single document summarization and shown promising results for summarizing news articles~\cite{Baralis2013Graphsum, Wei2015Gibberish}, emails~\cite{Carenini2008Summarizing, Paulus2017Deep}, product titles \cite{Sun2018Multi} etc. However, summarization methods for single (general) document are not applicable for customer reviews that are multiple documents written by various customers. 
Moreover, as customer reviews are typically short, informal text containing information about multiple aspects of products, it is hard to find the hidden topics among customer reviews and summarize them. 

Gernerally, summarization methods can be classified  into two categories: abstractive and extractive methods~\cite{Gambhir2017Recent, Pecar2018Towards}. Abstractive summarization aims to generate a short text summary by paraphrasing content of the original document. Several sentence compression methods have been proposed to comprise original sentences to create a summary by using a syntactic parser or a word graph \cite{Genest2010Text, Ffilippova2010Multi, Khan2015Framework}. However, it remains a difficult task due to the abstractive method typically involves many sophisticated nature language techniques such as meaning representation, content organization, sentence compression, paraphrasing etc. As such, the quality of the generated summary from an abstractive system is hard to control and present.  
Recently, many studies utilize neural networks that are based on encoder-decoder architecture for abstractive summarization~\cite{Rush2015Neural, Nayeem2018Abstractive, Cao2018Soft}. Still, the quality of the generated summary is the major concern, especially when it applies to customer reviews which contains noise, ungrammatical documents, and conflicting opinions.

Much more efforts have been focused on extractive summarization that aims to select salient parts of the original document such as sentence parts or whole sentences as the summarization of documents. As such, topic models and clustering method were introduced to find the documents that talk about the similar content/topic. Statistical features such as the position of sentences, positive and negative words, sentence length, etc. are used to select important sentences and words from the source text~\cite{Fattah2009GA, Abuobieda2012Text}. The encoder-decoder and attention mechanism also be applied for extractive summarization recently~\cite{Nallapati2017Summarunner}. However, all above methods focus on summarizaion of news articles, or emails, etc. There are still very limited number of studies focus on summarization of customer reviews~\cite{Zhan2009Gather, Yu2016Product, Amplayo2017Adaptable, Tan2017Sentence}. 

Zhan et al. (2009),  proposed an extractive summarizaiton method that is based on analysis of internal topic structure of product reviews and tested it on reviews collected from 8 products~\cite{Zhan2009Gather}. Yu et al. (2016) selected important sentences by analyzing their popularity and specificity~\cite{Yu2016Product}. For the methods proposed by Tan et al. (2017) and Amplayo et al. (2017), topic modeling methods are used~\cite{Amplayo2017Adaptable, Tan2017Sentence}. 
While thousand of reviews are used in above methods, they are heuristic methods due to the lack of groundtruth summaries. Summaries for less than 10 products or only the positive/negative rate are used to test their methods. As such, the critical source of summaries are missing for above methods.

Another work that could also be relevant to our work is opinion extraction from customer reviews. The opinion extraction methods differ from general customer review summarization as it focuses on summarizing selected sentences that only relevant to a manually designed topic. 
Hu et al. (2006) examined the review sentences and designed particular rules to detect product features among the source data and generate the summarization~\cite{Hu2006Opinion}. Ganesan et al. (2010) proposed a graph-based framework for generating summaries from review sentences  collected by using 51 queries~\cite{Ganesan2010Opiniosis}. Hu et al. (2017) proposed a sentence importance metric that is based on content and sentiment similarities for selecting important sentences~\cite{Hu2017Opinion}. Similarly, these methods are designed to learn opinions from only customer reviews rather than from both reviews and summaries.

Here we present a new topic modeling based summarization method with following main contributions. Firstly, we created a new Amazon-Cnet dataset with mapping between Amazon reviews and Cnet summary. Secondly, we provide a unified framework to segment review, cluster review sentiments into single document, model the review topics,and generate the summarization. Lastly, the experimental results and evaluation provide convincing  evidence  that the proposed method can be a useful tool for review summarization.

The rest of the article is organized as follows: Section 2 describes the complete framework of our method and detailed steps; Experimental settings, results and performance evaluation are presented and discussed in Section 3; followed by conclusion and future work in Section 4.

\section{Methods}
Our goal is to build a summary generator by analyzing both reviews and summaries. We first preprocess the raw text and segment reviews or summaries into individual sentiments where each of them contains information about only one aspect of product. As the sentiments are typically short, we thus combine sentiments talking about the same topic/aspect into a single document and apply topic modeling method to identify hidden topics among customer reviews and summaries. Next, we apply sentiment analysis method to those sentiments that belong to the same topic for distinguishing positive and negative sentiments. To generate the final summarization, a classifier also is introduced to distinguish the writing pattern of summaries and that of customer reviews. Finally, sentiments are selected to generate the summarization based on their topic probability, sentiment analysis score and writing pattern. The complete framework for our approach is shown in Fig. 1. The rest of this section provides details of each step.

\begin{figure}[!t]
\includegraphics[width = 3.3 in]{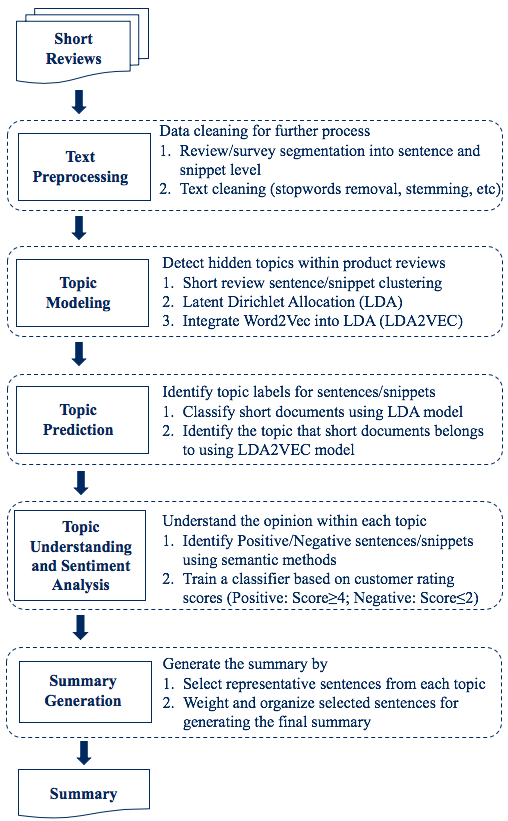}
\caption{Our framework for customer review summarization.}
\label{framework}
\end{figure}

\subsection{Text preprocessing}
A customer review typically contains information about multiple aspects of a product.
To obtain the information about individual aspect, we first break a review into sentences. 
As such, a customer review, \textit{review} $i$, is converted into a set of sentences $(s_i^1, .. s_i^m)$.
Similarly, we also split a summary, \textit{summary} $j$, into individual sentences $(s_j^1, .. s_j^n)$.

We note that customer review are written in informal and concise phrases. As such, the majority of sentences after parsing are very short (less than 8 words length). It is hard to learn the topic and sentiments from short documents directly. Notably, sentences contains the same noun. typically talking about the same aspect of the product. For example, ’The battery last one day long.’ and ‘It is pretty heavy due to the battery.’ all talk about ‘battery’ which appears in both sentences. We thus combine sentences that contain the same noun to create a longer document, $Doc_{noun}$, for further process. As a sentence may contain more than one noun, such sentence will appear in multiple combined documents. For the sentences that do not have any nouns, they will not be included in any combined documents.

For each sentence and combined document, the traditional preprocessing steps are also employed. We first substitute all contractions and specific terms, such as, e-mail, sd-card based on a manually build dictionary. For example, the word ‘e-mail’ will be converted to email. We also remove stop words~\cite{Nothman2018Stop} and standard suffixes using Porter stemmer~\cite{Porter1980}.

\begin{figure*}[htbp]
\includegraphics[width = 7 in]{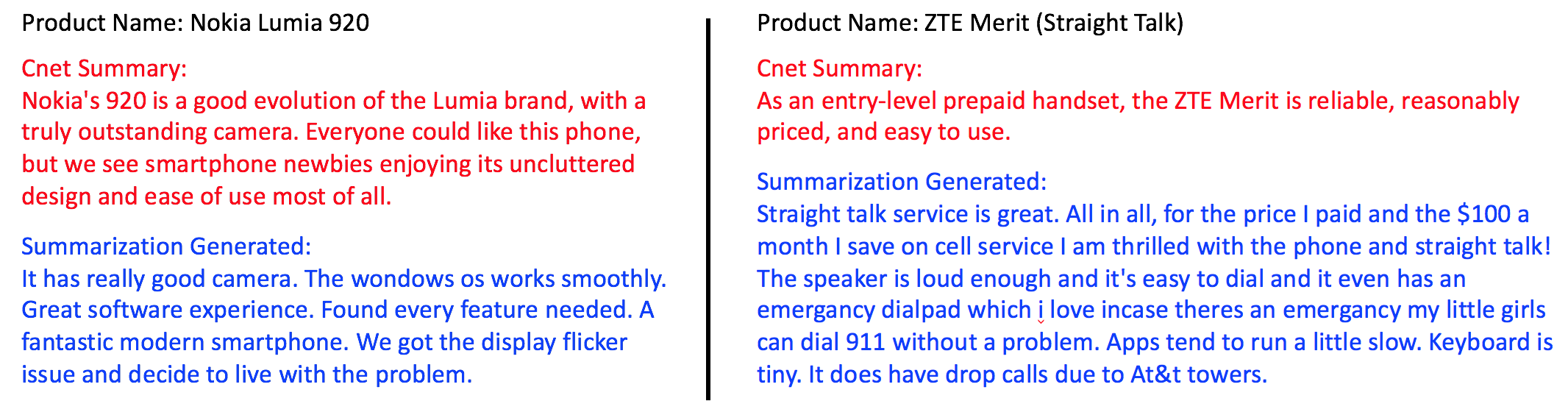}
\caption{Summarization examples generated by using our method.}
\label{summary_examples}
\end{figure*}

\subsection{Topic modeling on reviews and summaries}
As we mentioned before, customers reviews typically contain information about multiple aspects of products. Moreover, due to different user experience, the aspects reviewed by different customer could be also vary. Thus, identifying the hidden topics among customer reviews is important to generate the summarization. Regarding the summaries, as they are written by people typically with expertise and focus on only product itself. The topics covered by customer reviews are quite different from that by summaries. For example, the shipping experience which is an important topic among customer reviews, typically will not be mentioned in Cnet summaries. We thus identify the hidden topics among both customer reviews and summaries. 

In this work, we use LDA to identify hidden topics among combined documents of customer reviews and of summaries, which is a generative statistical model that has been widely used for topic modeling. To do that, we use the implementation from scikit-learn~\cite{Hoffman2010Online}. The model parameters are learnt iteratively for different number of topics, $K$, where $K$ ranges from 5 to 40, and the log-likelihood and perplexity are calculated for each value of $K$. To determine the optimal number of topics, we identify the $K$ value that maximizes log-likelihood and minimize the perplexity. Two models LDAreview and LDAsummary are trained based on Amazon customer reviews and Cnet summaries respectively.

To identify the review sentences that talk about the same topic, we predict the topic label for each review sentence using LDAreview and LDAsummary, respectively. We note review sentences after the parsing step may be too short to be classified. Therefore, the sentence which obtains all zero prediction for all topics will be discard. 
Review sentences then will be grouped into sets of topics along with their probability scores, $P_{s_i^j}^t$, indicating how likely the sentence $s_i^j$ belongs to the particular topic $t$.

\subsection{Topic understanding and sentiment analysis}
While we have split review sentences into sets of topics, many sentences that belongs to the same topic may express conflicting opinions. For example, sentences ’The screen has great resolution.’ and ‘I hope I bought larger screen’ could be assigned with the same topic label, while they totally express opposite opinions. To identify the hidden opinions among each topic, we apply sentiment analysis to sentences belonging to the same set.

To do sentiment analysis, we employ VADER (Valence Aware Dictionary for sEntiment Reasoning) which is a rule-based model utilizing lexical features and rules that embody grammatical and syntactical conventions \cite{Hutto2014Vader}. As VADER is build upon analyzing social media text snippets collected from twitters that have similar writing patterns with review data, we believe it is well-fitted for review sentiment analysis.

Thus, given a review sentence $s_i^j$, we can obtain a positive sentiment score, $PS_i^j$, and a negative sentiment score, neg $PS_i^j$ by using VADER. The opinion of sentence $s_i^j$ then can be determined by the label of the maximum value of  $PS_i^j$ and $PS_i^j$. 

\subsection{Summary generation}
Customers typically talk about multiple aspects of the product in their reviews. To generate the final summary, we first identify the $k$ most salient topics that are covered by customer reviews. In our work, the $k$ is set to 5. 
To do that, we check the number of sentences that in each topic set and pick the $k$ most salient topics for generating the summary.

As we mentioned, conflicting opinions could appear in the same topic. To represent the overall opinion of a topic, we also select the most popular attitude in the most salient topics. For example, when the number of positive sentences is higher than that of sentences labeled as negative, we believe the topic is positive and the opinion score for each sentence obtained by sentiment analysis, $OP_i^j = PS_i^j$. Otherwise, the topic is negative and the opinion score for each sentence, $OP_i^j = NS_i^j$. 

Notably, summaries are typically written in a different style from customer reviews. Therefore, the writing style is one of the most important factors for ranking sentences.
Unlike existing work that generate the summary just based on the analysis of reviews. We build a classifier to distinguish writing patterns for summaries and that for customer reviews. To do that, we create a set of summary sentences and a set of review sentences. We note the imbalance issue between the summary sentence set and the review sentence set. We use meta learning~\cite{Chan1998Toward} where the majority class is split into multiple subsets, each of which is of similar size to the minority class, to train a base-classifier. The final classifier is build upon the decision made from all base classifiers. By using this classifier, we can obtain a summary likelihood for each sentence, $SL_i^j$.

The summarization of customer reviews is a set of most important/informative/representative sentences that are selected from the $k$ most salient topics. All above factors are very critical to determine the importance of the sentence:
\begin{itemize}
\item The probability that the sentence belong to current topic: $P_{s_i^j}^t$
\item The opinion score obtained from sentiment analysis: $OP_i^j$.
\item The summary likelihood: $SL_i^j$.
\end{itemize}

To select the most important sentence within each topic sentence set, we calculate the importance score for each sentence as follows:
\[Score(S_i^j) = (P_{s_i^j}^t + OP_i^j) * SL_i^j\]

After ranking the sentences within each topic, sentences with the highest score within their corresponding topics are selected as the final summary. 

\section{Experiments and results}
\subsection{Experimental settings}
To evaluate our method, we compare the performance of our system to that of two state-of-the-art systems, TextRank~\cite{Mihalcea2004Textrank}, Opinosis~\cite{Ganesan2010Opiniosis}, Biclique~\cite{Muhammad2016biclique}, ILPSumm~\cite{Banerjee2015ILPsumm}, and ParaFuse\_doc~\cite{Nayeem2018Abstractive}. 
Automatic evaluation measures like ROUGE and its modifications~\cite{Lin2004Rouge} are used to evaluate our performance. 

\subsection{Datasets}
As there is no relevant dataset available online, we build our own, Amazon-Cnet dataset. To do that, we first select 2000 cell phone products that associated with more than 10 customer reviews at random from the Cell Phones and Accessories category in Amazon Review Dataset~\cite{He2016Ups}. Cnet \footnote[1]{www.cnet.com} is one of the most popular website providing professional reviews for electronics, such as cell phones. We thus manually crawl the summary from Cnet webpage as ground-truth summaries for those cell phones. At last, 1028 products have their Cnet review webpage and be included in Amazon-Cnet datsset. Table 1 provides statistics of our datasets.

\begin{table}[htbp]
\centering
\caption{Statistics for our Amazon-Cnet dataset used in our experiments. The summaries and customers reviews are for 1028 products.}
\label{table_dataset}
\begin{tabular}{l | c | c } 
\toprule
\textbf{}   & \textbf{The number} & \textbf{\# of sentences} \\
\midrule
Summaries  & 1028   & 1385\\
Reviews & 66129 &  362965 \\
\bottomrule
\end{tabular}
\end{table}

\subsection{Results}

Our method attained 15.43\% over the Amazon-Cnet dataset.
We note that the ROUGE-1 score obtained by our method is still much lower than other results reported over datasets like Opinions, given Amazon-Cnet datsset is more challenging and practical for real-world usage. 
By looking at the dataset, we found that the reviews typically talk about details of a product, such as the resolution of the screens, the loudness of the speakerphone. 
In contrast, the summaries typically talk about the high level characteristics of the product, such as the smooth of the mobile system.  
We also found that the topics within customer reviews usually are not interests of summarization. For example, the customer service, which is a hot topic among reviews, is not an interest for summarization.
Unlike other methods generate summarization by analyze the reviews only, our method also consider the topics among summaries therefore out-performance all other methods.
Figure~\ref{summary_examples} shows two summarization examples that are generated by using our method. 

This is our first preliminary work to integrate the summary information for customer review summarizations. As such, there is still much room for improvements. We are going to collect more online review and sumamrization pairs to train a more comprehensive model. 
We also note that the user reviews of a product could span a wide range of time period. However, the summary of a product is typically posted at very early stage when the product been released. For example, the Cnet summary for product 'ZTE Merit' is posted on Aug 2012, and the latest customer review is posted on Mar 2016. We can believe that the opinion reported 4 years after the first product release may not be helpful for hte summariaation.
As such, we are going to also investigate the time series issue among customer reviews for creating their summarization.

\section{Conclusion}
 
We present a new method for customer review summarization utilizing both customer reviews and summaries. Unlike other methods that only consider customer reviews, we identify hidden topics among both customer reviews and summaries. 
Sentiment analysis is employed to distinguish positive and negative opinions among each detected topic. 
A classifier is also introduced to distinguish the writing pattern of summaries and that of customer reviews. 
Finally, sentiments are selected to generate the summarization based on their topic relevance, sentiment analysis score and the writing pattern. A new dataset comprising product reviews and summaries about 1028 products are collected from Amazon and CNET. Experimental results show the effectiveness of our method. 

Sevearl challenging issues remain as future work. We are going to collect more data to train a more comprehensive model for review summarization. We shall also investigate the time series among customer reviews for generating summarization.

\end{document}